\documentclass[letterpaper, 10 pt, conference]{ieeeconf}  % Comment this line out if you need a4paper

\IEEEoverridecommandlockouts                              % This command is only needed if 
\usepackage[T1]{fontenc} % Ensures proper font encoding
\usepackage{lmodern} % Loads the Latin Modern font which is a more modern version of the typewriter font
\usepackage{times} % Uses Times font for text which is smoother than typewriter fonts

\usepackage{graphics} % for pdf, bitmapped graphics files
\usepackage{epsfig} % for postscript graphics files
\usepackage{mathptmx} % assumes new font selection scheme installed
\usepackage{times} % assumes new font selection scheme installed
\usepackage{amsmath} % assumes amsmath package installed
\usepackage{amssymb}  % assumes amsmath package installed
\usepackage[]{hyperref}
\usepackage[]{cite}

\usepackage{tabularx}
\usepackage{booktabs}
\usepackage{multicol}
\usepackage{multirow}
\usepackage{floatrow}
\usepackage{hyperref}
\usepackage{graphicx}
\usepackage{wrapfig}
\usepackage{tikz}
\usetikzlibrary{tikzmark}

\usepackage{xcolor}

\usepackage{listings}

\usepackage{algorithm}
\usepackage{algpseudocode}

\definecolor{backcolourwhite}{rgb}{1,1,1}
\definecolor{codegreen}{rgb}{0,0.6,0}
\definecolor{codegray}{rgb}{0.5,0.5,0.5}
\definecolor{codepurple}{rgb}{0.58,0,0.82}
\definecolor{backcolour}{rgb}{0.95,0.95,0.92}
\definecolor{light-gray}{gray}{0.98}
\definecolor{blue}{rgb}{0.,0.,0.90}
\definecolor{light-blue}{rgb}{0.67,0.85,0.90}

\lstdefinestyle{mystyle}{
    backgroundcolor=\color{light-gray},   
    basicstyle=\ttfamily\footnotesize,%\setstretch{.5},
    commentstyle=\ttfamily\color{codegreen},
    keywordstyle=\ttfamily\color{magenta},
    breakatwhitespace=false,         
    breaklines=false,                 
    captionpos=b,                    
    keepspaces=true,                                     
    numbersep=5pt,                  
    showspaces=false,                
    showstringspaces=false,
    showtabs=false,                  
    tabsize=2,
    frame=single,
    rulecolor=\color{black!40},
    escapechar={|}, 
    language=Python
}

\usepackage[tableposition=top]{caption}

\title{\LARGE \bf
Lifelong Robot Library Learning:  Bootstrapping Composable and Generalizable Skills for Embodied Control with Language Models
}

\author{Georgios Tziafas$^{1}$ and Hamidreza Kasaei$^{1}$% <-this % stops a space
\thanks{$^{1}$Department of Artificial Intelligence,
        University of Groningen, the Netherlands,
        {\tt\small \{g.t.tziafas,hamidreza.kasaei\}@rug.nl}}%
}

\begin{document}

\maketitle
\thispagestyle{empty}
\pagestyle{empty}

%%%%%%%%%%%%%%%%%%%%%%%%%%%%%%%%%%%%%%%%%%%%%%%%%%%%%%%%%%%%%%%%%%%%%%%%%%%%%%%%
\begin{abstract}
Large Language Models (LLMs) have emerged as a new paradigm for embodied reasoning and control, most recently by generating robot policy code that utilizes a custom library of vision and control primitive skills.
However, prior arts fix their skills library and steer the LLM with carefully hand-crafted prompt engineering, limiting the agent to a stationary range of addressable tasks. 
In this work, we introduce \texttt{LRLL}, an LLM-based \textbf{li}felong \textbf{l}earning agent that continuously grows the robot skill library to tackle \textbf{m}anipulation tasks of ever-growing complexity.
\texttt{LRLL} achieves this with four novel contributions: 1) a soft memory module that allows dynamic storage and retrieval of past experiences to serve as context, 2) a self-guided exploration policy that proposes new tasks in simulation, 3) a skill abstractor that distills recent experiences into new library skills,  and 4) a lifelong learning algorithm for enabling human users to bootstrap new skills with minimal online interaction. 
\texttt{LRLL} continuously transfers knowledge from the memory to the library, building composable, general and interpretable policies, while bypassing gradient-based optimization, thus relieving the
learner from catastrophic forgetting. Empirical evaluation in a simulated tabletop environment shows that LRLL outperforms end-to-end and vanilla LLM approaches in the lifelong setup while learning skills that are transferable to the real world. Project material will become available at the webpage \href{https://gtziafas.github.io/LRLL_project/}{\textcolor{codepurple}{https://gtziafas.github.io/LRLL\_project/}}.
\end{abstract}

%%%%%%%%%%%%%%%%%%%%%%%%%%%%%%%%%%%%%%%%%%%%%%%%%%%%%%%%%%%%%%%%%%%%%%%%%%%%%%%%
\section{INTRODUCTION}
Building interactive agents that can continuously develop new skills and adapt to new scenarios remains a challenging frontier in robotics \cite{Kolve2017AI2THORAI,Ehsani2021ManipulaTHORAF,Savva2019HabitatAP}.
Such an agent should be able to interface natural language, percepts and actions in order to form policies that are reusable and expandable in an open-ended fashion \cite{Tellex2020RobotsTU}.
Recent advances in end-to-end robot learning \cite{RT2VM, PaLMEAE, VIMAGR, InteractiveLT} learn capable multimodal policies but require copious amounts of data, which are very hard to scale in the robotics domain. 
Further, their reliance in gradient-based optimization hinders their applicability in a lifelong setup, due to the effect of catastrophic forgetting \cite{Lesort2019ContinualLF, Parisi2018ContinualLL, Wang2023ACS}.

Meanwhile, an emerging paradigm has been to leverage the code-writing capabilities of modern LLMs \cite{Brown2020LanguageMA,Touvron2023Llama2O,Chowdhery2022PaLMSL} for synthesizing executable robot policy code from natural language \cite{CodeAP,Instruct2ActMM,ProgPromptGS,ChatGPTFR, VoxPoserC3}.
% LLM-based agents have been employed in a broad range of domains, including decision-making in games [] or NLP [], embodied task planning [], navigation [], as well as manipulation [].
In such a setup, vision and action skills are implemented as modules (either learned or scripted) in a first-party API.
This allows the LLM to compose them arbitrarily in combination with classic programming structures (control flow, recursion etc.) and third-party Python APIs (e.g. \texttt{numpy}) in order to ground visual observation, perform low-level reasoning, and provide parameters for control primitives.
This system bypasses model finetuning, instead relying on careful prompt design and in-context examples to steer the LLM and aid generalization. 
However, the choice of the skills library and prompt examples remains a design choice that limits the span of tasks that the agent can tackle, and require an expert to continuously adapt the library and prompts to the LLM.
In this work, we wish to address such limitations by proposing $\textit{LRLL}$, a \textbf{L}ifelong \textbf{R}obot \textbf{L}ibrary \textbf{L}earning agent.
\textit{LRLL} learns hierarchical and generalizable skills across time spans, while staying within the regime of in-context learning and bringing non-expert human users in-the-loop.
\begin{figure}[!t]
    \includegraphics[width=9cm,height=40cm,keepaspectratio]{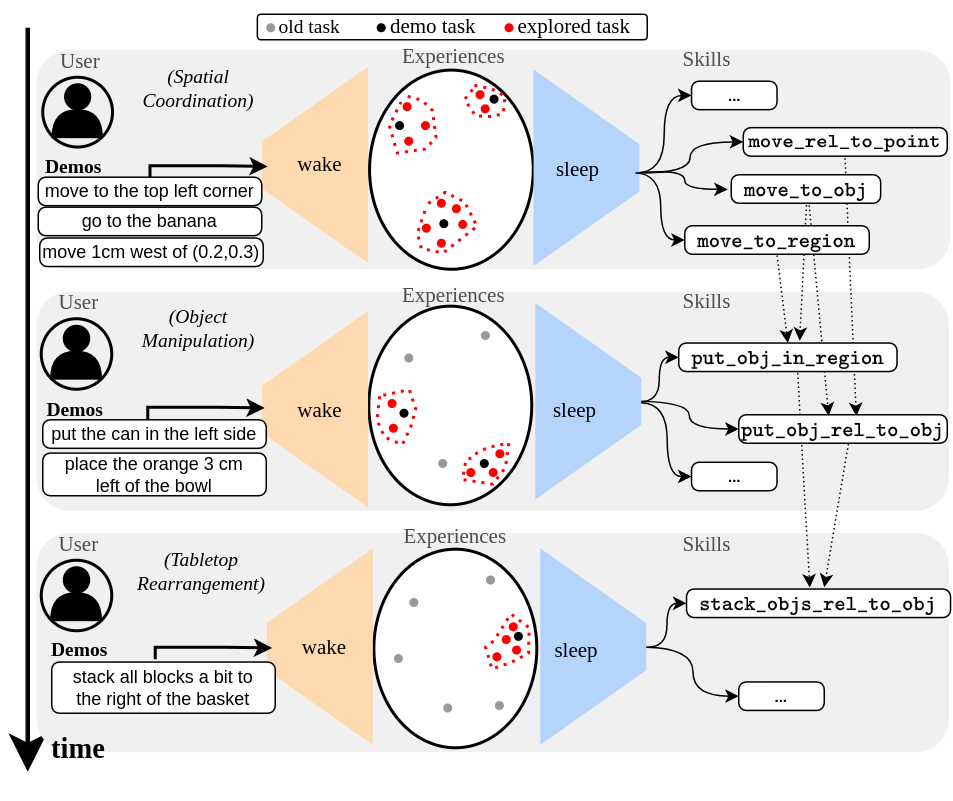} 
\caption{\footnotesize
Wake-sleep library learning from human guidance.
}
\label{fig:f1}
    \vspace{-5mm}
\end{figure}

Our learning algorithm is inspired by \textit{wake-sleep optimization} \cite{Hinton1995TheA} and its adaptation for library learning \cite{DreamCoderBI}.
Learning takes place in cycles, each with two distinct phases: a) a \textit{wake} phase, during which the agent interacts with its environment and users in order to grow its experiences, and b) a \textit{sleep} phase, during which the agent reflects on its experiences in order to expand its capabilities.
Accumulated experiences are distilled into skills throughout the learning cycles, therefore allowing complex tasks to be expressed as programs composed of simpler skills (see Fig~\ref{fig:f1}). 
The human acts as a teacher, introducing a few demonstrations and hints to the agent during the wake phase.
We assume the teacher follows a curriculum approach, where in each cycle the objective tasks can be built out of the learner's current repertoire of skills.

To stay within in-context learning, we use a frozen LLM to generate the policy code, and design our algorithm as an interchange between two modules: a) an \textit{experience memory}, where the agent can store and retrieve past instruction-code pairs based on similarity search, in order to feed context to the LLM (i.e. prompt retrieval), and b) a \textit{skill library}, which comprises the collection of API calls the LLM can generate code from.
At each cycle, the wake phase populates the memory with new instruction-code pairs, generated from an LLM-based exploration module that proposes and self-verifies new tasks in a simulator.
During the sleep phase, the accumulated experiences are distilled into new skills via an LLM-based abstraction module. 
The new skills are appended to the library and the wake phase is replayed with refactored experiences, in order to compress the memory.
This leads to a continual transfer of knowledge from human guidance, via exploration in simulation, to the library, without utilising gradients in any part of the process.
% Finally, to reduce the dependency in human feedback and enhance the quality of skill abstraction and evaluation, we additionally leverage the LLM to propose new task variations and self-test the recently acquired skills at the end of both stages.
% Such agent should possess the ability to reason about percepts and actions to form reusable skills, as well as invent new skills to adress tasks in an open-ended fashion.
% Modern works in the field of lifelong learning [] focus on combating catastrophic forgetting within the regime of gradient updates [], but are mostly limited to vision domains [], still far from application in embodied interactive agents.
% On the other hand, an emerging paradigm has been to utilize the Internet-scale world knowledge and common-sense reasoning capabilities of modern LLMs [] to synthesize action plans from natural language commands. 

To apply our idea in robotics, we design a four-stage curriculum and prompt our agent to acquire a broad range of skills, including precise visual-spatial reasoning and long-horizon tabletop rearrangement.
Empirically, we show that \textit{LRLL} can automatically build a library of hierarchical, generalizable and interpretable skills, while outperforming end-to-end and stationary LLM baselines.
We further perform ablations to demonstrate the effectiveness of each proposed component and explore design options. 
Finally, we illustrate that our algorithm can be transferred to a real robot for dual-arm tabletop rearrangement tasks, without any further adaptation.
In summary, our key contributions are the following: a) \textit{LRLL}, an LLM-based agent that can generate policy code, explore tasks in simulation, and expand its skillset over time, b) a formal recipe for enabling humans to bootstrap desired robot skills with minimal intervention, and c) extensive comparisons, ablation studies and hardware demonstrations that evaluate the effectiveness of each proposed component, assess overall generalization capabilities and test sim-to-real transferrability.

\section{RELATED WORK}
\textbf{Language to Action} Natural language has a long-standing history for controlling robots \cite{Tellex2020RobotsTU}, serving both as a natural interface for human-robot interaction \cite{TellMeDave, holisticNlgrasp1, INGRESS, INVIGORATEIV}, as well as a generalizable intermediate representation \cite{BCZZT, CLIPortWA, VIMAGR}.
Approaches range from semantic parsing \cite{Thomason2015LearningTI, ProgrammaticallyGC, Tziafas2022EnhancingIA}, planning \cite{ZSP, SayCan, InnerME, SocraticMC}, reinforcement \cite{Luketina2019ASO, PixL2RGR, Jiang2019LanguageAA}, imitation \cite{BCZZT,CLIPortWA,VIMAGR,Stepputtis2020LanguageConditionedIL,InteractiveLT}, and model-based \cite{Nair2021LearningLR, Andreas2017LearningWL, Sharma2022CorrectingRP} learning to more recent large-scale end-to-end multimodal instruction-following \cite{RT2VM, EmbodiedGPTVP}. 
While end-to-end policies are becoming more capable, they require prohibitive amounts of offline data or environment interactions. 
Further, their lifelong learning potential is limited by the effect of catastrophic forgetting \cite{Lesort2019ContinualLF, Parisi2018ContinualLL, Wang2023ACS}. 
In contrast, in this work we focus on a gradient-free approach where low-level actions are implemented as control primitives, out of which an LLM continuously builds more complex skills via few-shot human demonstration and interactive exploration.

\noindent \textbf{LLMs for Robot Control} More similar to this work, an emerging body of methods is chaining LLMs with external models \cite{CLIP, ViLD, MDETRM} in order to propose grounded plans that sequence high-level actions \cite{ZSP, SayCan, InnerME, SocraticMC}.
This method invests on the current capabilities of LLMs for multi-step reasoning using external modules as tools \cite{ReActSR}, without adittional finetuning.
Recent works \cite{CodeAP, ProgPromptGS, ChatGPTFR, Instruct2ActMM} replace high-level actions with a library of primitives, and use the LLM to generate Python code that grounds the visual scene and parameterizes the primitives.
This allows more complex policy logic than sequences of actions and offers more precise spatial grounding \cite{VoxPoserC3} and reasoning \cite{CodeAP}.
However, such works are stationary systems that do not further extend their library, and require manual prompt engineering to be applied in a general setup. 
\textit{Code-as-Policies (CaP)} \cite{CodeAP} demonstrates sparks of non-stationarity by letting the LLM recursively define unseen functions, but does not do so in a controlled, reusable fashion and does not consider human guidance.
In this work, we wish to extend a \textit{CaP}-like system to incorporate past interactions as a self-prompting mechanism and systematically use the LLM function generator to expand the skill library over time.

\noindent \textbf{Memory and Context Retrieval} Retrieval-augmented LLMs are a trending direction in NLP research \cite{Shi2023REPLUGRB, Jiang2023ActiveRA, Nakano2021WebGPTBQ}, mostly as a means to reduce LLM hallucinations. 
In the robotics and embodied AI space, several works retrieve the most similar task-code pairs from memory based on similarity search and use them to prompt the LLM \cite{ProgPromptGS, ZSP}, but do so in a stationary fashion.
Recent works \cite{GenerativeAI, VoyagerAO, AssistGPTAG} expand the memory based on interactions, but there is no refinement of the base skills.
In our work, we use the memory for LLM prompt retrieval, but progressively distill similar experiences to new skills that refactor and compress the memory.

\noindent \textbf{Language-guided Skill Acquisition} Iteratively refining robot skills with language feedback has been explored in the past with external parsers \cite{Broad2017RealtimeNL} and end-to-end language-conditioned policies \cite{Cui2023NoTT, Sharma2022CorrectingRP, Bucker2022ReshapingRT,
Bucker2022LATTELT}, but rely on either domain knowledge or extensive demonstration datasets, and therefore lack scalability.
More recently, \cite{LanguageTR} leveraged LLMs with multi-step human feedback to generate reward functions that will train policies with model-predictive control.
The concurrent work \cite{ScalingUA} utilizes an LLM to generate synthetic experiences in simulation together with success conditions, and distills the successful trials into a language-conditioned policy with behavioral cloning.
Both ideas are conceptually close to our work, exploiting the LLM to generate information that will train a policy, but employ traditional approaches for learning the policy and hence struggle with the lifelong learning setup.
Instead, \textit{LRLL} leverages control primitives and an LLM to express the policy itself, and poses skill acquisition as library learning on top of the primitives. This allows complex skills to form from simpler skills over time, while retaining interpretability and bypassing training policies from scratch.

\section{METHOD}
\begin{figure*}[!t]
    \includegraphics[width=1\textwidth]{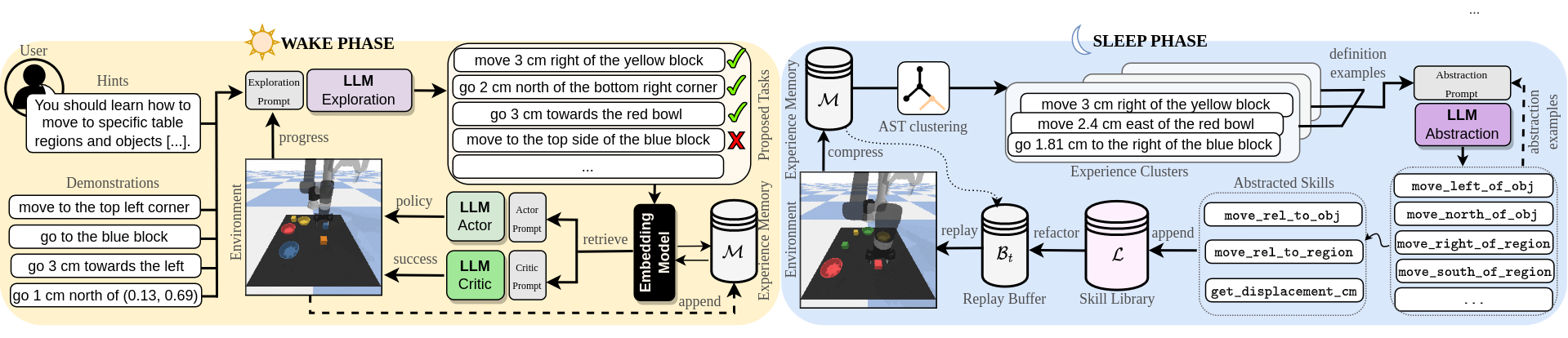} 
\caption{\footnotesize
Overview of an \textit{LRLL} learning cycle. At the beginning of the wake phase, a human user provides demonstrations and hints, out of which an LLM-based \textit{exploration} module proposes tasks to complete, while an LLM-based \textit{actor-critic} agent interacts with the environment to execute and verify tasks. During sleep, the experiences are clustered according to their code's abstract syntax trees (AST) and distilled into new skills with an LLM \textit{abstractor}. The new skills refactor the acquired experiences, which are replayed in the environment in order to compress the memory. We note that for brevity purposes, we omit showing the actor-critic modules during replay at the sleep phase. We illustrate examples from our first curriculum cycle (\textit{spatial coordination}).
}
\label{fig:schematic}
    \vspace{-3mm}
\end{figure*}

In this section, we provide an overview of the proposed algorithm (Sec.~\ref{method:overview}) and its components (Sec.~\ref{method:system}), and describe details for achieving policy and success generation, exploration and skill abstraction with LLMs (Sec.~\ref{method:llm}).

\subsection{LRLL Overview}
\label{method:overview}
% \begin{algorithm}[!b]
% % \caption{OWG}\label{alg:cap}
% \caption{LRLL Wake-Sleep Cycle}\label{alg:cap}
% \begin{algorithmic}
% \Require Memory $M_{t}$, Library $L_{t}$, Demos $X_t = <l,a,r>_t$, Hints $G_t$, wake iterations M
% \State \# \textbf{Wake Phase}
% \State $B \gets \varnothing $ \Comment Initialize replay buffer
% \For {$m = \{1, ..., M\}$} \quad 
%     \State $\tilde{l}_{1:K} \gets$ \texttt{LLMExplore}($X_t$,$G_t$,$B_t$) \Comment{Propose tasks}
%     \For {$\tilde{l}_i = \{\tilde{l}_1, \dots, \tilde{l}_K \}$} 
%         \State Sample random initial simulator state $s_{0,i}$
%         \State $\tilde{a}_i, \tilde{r}_i \gets 
%         \texttt{LLMActorCritic}(\tilde{l}_i, M_t', s_{0,i})$
%         \State Execute program $\tilde{a}_i$ and evaluate success $\tilde{r}_i$
%     \EndFor
%         \State $M_t' \gets M_t \cup \left \{ <\tilde{l}_i,\tilde{a}_i,\tilde{r}_i>_{i=1}^K \; \mid \; \tilde{r}_i=1  \right \}$ 
%         \State $B_t \gets B_t \cup \left \{ <s_{0,i}, \tilde{l}_i,\tilde{a}_i,\tilde{r}_i>_{i=1}^K \; \mid \; \tilde{r}_i=1  \right \}$ 
% \EndFor
% \State 
% \State  \# \textbf{Sleep Phase}
% \State $C_{1:N} \gets \texttt{ASTParse}(M_t')$ \Comment{Cluster experiences}
% \For {$<l,a,r>_n \; \in \; C_{1:N}$}
%     \State $F_n$ = \texttt{LLMAbstract}($<l, a, r>_n$) \Comment {Propose skills}
% \EndFor
% \State $L_{t+1} \gets L_t \; \cup \left \{F_{1:N}  \right \}$ 

% \end{algorithmic}
% \end{algorithm}

Our framework (see Fig.~\ref{fig:schematic}) receives at each cycle $t$ a set of $N_t$ language demonstrations $X_t$ that contain instruction-policy-success tuples for specific tasks: $X_t = \left \langle l^{(i)}, a^{(i)}, r^{(i)} \right \rangle_{i=1}^{N_t}$.
The demos are appended to the agent's memory $M_t$.
The policy code $a^{(i)}$ is factorized based on the current library skills $L_t$. 
The success code $r^{(i)}$ describes how the agent can use privileged simulation data to verify the success of a given instruction (e.g. check contact, object poses etc.).
The output is a set of formal skills, expressed as Python functions, which can be directly used for real robot deployment.
This process is repeated for an open-ended number of cycles, enabling the agent to expand its library in lifelong fashion.
Each cycle consists of two phases:

\noindent \textbf{Wake Phase} During the wake phase, the agent interacts with its environment in order to grow its proficiency in solving more tasks. 
We use an LLM to iteratively propose new task instructions $l_{1:K} = \texttt{LLMExplore}(X_t, G_t, M_t)$ based on the input demos $X_t$, the current memory state $M_t$ and some general hints $G_t$. 
The proposed tasks are executed and verified in the simulator using another LLM's generated policy and success code respectively: ${a_k}, {r_k} = \texttt{LLMActorCritic}(~{l}_k, s_{0,k}, M_t)$, where $s_{0,k}$ the initial simulator state of proposal $k$.
Successful  tuples are appended to the experience memory $M_t$, which helps the agent to recover more relevant context throughout exploration.
The process is repeated until an iteration threshold is met, or the LLM decides that it has completed all objectives denoted in the hints. The acquired experiences are stored in a replay buffer $B_t = \left \{ <s_{0,k}, l_k, a_k, r_k> | \; r_k=1 \right \}$.

\noindent \textbf{Sleep Phase} During sleep, the agent reflects on its acquired experiences to compose new skills. To achieve this, we first represent each experience as an abstract syntax tree of its policy code.
Experiences are clustered such that codes that have the same tree structure modulo variable and constant names are grouped together for a total of M clusters $C_{1:M}$. 
We then feed each cluster into an LLM that uses the experiences as examples to define new functions that will update the library $L_{t+1} = L_t \cup \{\texttt{LLMAbstract}(C_m)\}_{m=1}^M$. 
The demo policies are refactored based on the proposed functions $\tilde{X}^{(i)}=\texttt{Refactor}(X^{(i)}, L_{t+1})$, and the wake phase is replayed from scratch, starting only with the refactored demos. 
When the agent fails in a previously succeeded task, its policy-success is also refactored and appended to the memory:
$M_{t+1} = M_t \cup \left \{ \tilde{X}^{(i)} \in B_t \; | \; r^{(i)}=0 \right \}$
This process ensures that by the end of the cycle, the memory will contain the minimum number of experiences needed to replicate the performance of the wake phase.

\subsection{System Components}
\label{method:system}

\noindent \textbf{Initial Vision \& Action Primitives} Following previous works \cite{CodeAP, Instruct2ActMM}, we employ frozen pretrained vision-language models for zero-shot vision-language grounding
In particular, we use MDETR \cite{MDETRM} for referring expression grounding and CLIP \cite{CLIP} for open-vocabulary classification.
For control, as we wish to demonstrate the capability of LRLL to progressively build complex skills from simpler ones, we start with the most basic primitives: moving the arm to a certain pose and opening / closing the gripper. 
Motion planning is performed via inverse kinematics from end-effector space.

% \noindent \textbf{LLM} We use LLMs for: a) \textit{policy generation}, i.e. producing action code, b) \textit{success generation}, i.e. producing a success function, c) \textit{exploration}, i.e. interacting with a simulator to propose and complete new tasks based on human guidance, and c) \textit{skill abstraction}, i.e. compressing explored task experiences into new skills.

\noindent \textbf{Experience Memory \& Retrieval} Agent experiences are formalized as tuples of task instruction, action and success code, either provided by the human teacher or ``imagined" by the LLM exploration module $X^{(i)} = <l^{(i)}, a^{(i)}, r^{(i)}>$. When appended to the memory, each experience is indexed by its instruction embedding, provided by an encoder-based LM \cite{Reimers2019SentenceBERTSE, OpenAIEmbeddings} $\mathbf{z}_i = F^{LM}(l^{(i)})$. 
In order to retrieve experiences to serve as prompts, the query instruction $q$ is embedded by the same model $\mathbf{z}_q$ and the experiences of the top-\textit{k} most similar instructions based on maximum marginal relevance search \cite{Carbonell1998TheUO} are returned:
\begin{equation*}
    \texttt{argmax}_{i \in M_t \mid  S } \; \left [ \lambda (\texttt{cos}(\textbf{z}_i, \textbf{z}_q) - (1-\lambda) \texttt{max}_{j \in S} \; \texttt{cos} (\textbf{z}_i, \textbf{z}_j)) \right ]
\end{equation*}
where $S$ the set of already selected retrievals, $\texttt{cos}$ the cosine distance metric and $\lambda$ a diversification hyper-parameter.
This rule is applied $k$ times to retrieve diverse experiences.

\noindent \textbf{Skill Library} Each agent skill corresponds to a function, implemented as a Python API.
% The library can be queried for skill information, such as the skill name, description (docstring), code, as well as a dependency tree of required skills. This allows the library to render prompts with skill information as context to the LLMs.
The library maintains skill information such as their names and descriptions, and is able to trace skill dependencies from a given code snippet.
Before learning begins, the library is initialized with the initial primitives.
A wrapper around the library and the agent converts the newly acquired skills into API modules that are executable in a robot simulator.

\noindent \textbf{Replay Buffer} The replay buffer is a replica of the experience memory but only for the explored experiences of the current cycle. Additionally, for each experience, the simulator states are saved. The replay buffer is reset at the beginning of each new cycle.

\subsection{LLM Prompts}
\label{method:llm}
% \subsection{Actor-Critic: Policy and Success Code Generation}
% \label{method:actor-critic}
% In this work, we move from static prompts fed to LLMs to sequences of dynamic prompt templates and LLM calls chained together, enabling LLM outputs to behave as prompt fields in the input of another LLM call. 
We implement three LLM modules: 
% Three modules are implemented: 1) policy and success code generation \textit{(Actor-Critic)}, 2) exploration of new task variations and compositions \textit{(Exploration)}, 3) Python function definitions from a set of example code snippets (\textit{Abstraction}). 

\noindent \textbf{Actor-Critic} The actor-critic comprises of two parallel LLM calls, one for policy and one for success code generation. Both prompt templates are instantiated after retrieving experiences from the memory for a given query, and follow the same general structure:
\begin{itemize}
    \item A comment indicating a \textbf{general purpose} of the code (e.g. \textit{"\#\#\# Python robot control script"}).
    \item Information about the \textbf{API}, aiming to present the building blocks for code generation. For the actor, all modules are extracted from the retrieved examples' policies and rendered as import statements \cite{CodeAP}. For the critic, a fixed code snippet describing simulator utilities via docstrings is provided \cite{ScalingUA}.
    \item A sequence of task-code pairs from the \textbf{retrieved experiences}. Each pair is rendered as a comment of the task description followed by a policy or success code snippet, for the actor and critic respectively.
    \item Throughout demonstration code, \textbf{chain-of-thoughts} \cite{ChainOT} are provided as in-line comments to guide the LLM's reasoning before producing a next line, which is especially useful for explaining perspective conventions (e.g. \textit{"left"} correspond to x-axis).
\end{itemize}
For inference, the query is appended to the prompt as a comment and the LLM fills the corresponding policy or success code.
We find that this code-based completion format works robustly also for chat-based LLMs \cite{OpenAIChatGPT}.

% \subsection{Exploration: Collecting and Self-Verifying Experiences} 
% \label{method:explore}
\noindent \textbf{Exploration} The exploration module proposes the next tasks to complete in the simulator.
The goal is to use the demos as guidance and introduce both task \textit{variations}, i.e. alter concepts present in the instruction (e.g. color, spatial direction etc.), as well as task \textit{compositions}, i.e. combinations of concepts present in the demos (e.g. desired destinations for placing objects). The prompt contains:
\begin{itemize}
    \item  A system message that includes \textbf{general directives} that condition the LLM for the task, encourages diverse responses and provides the required response format.
    \item \textbf{Hints} provided by the teacher, aimed to provide objectives of a specific cycle.
    \item Information about the current \textbf{state}, represented as a list of appearing object names.
    \item A list of completed and failed tasks so far, reflecting the agent's current \textbf{progress} towards completing all objectives mentioned in the hints. The completed tasks are initialized with the provided demos.
    \item A set of two \textbf{exemplar} generations, containing input demo-hints and output task proposals. We provide one manual and set one more exemplar from the LLM's first response in the previous cycle.    
\end{itemize}
Before proposing tasks, we ask the LLM to reason about its proposals \cite{VoyagerAO}, which significantly helps in responding better to the provided hints. 
% Proposed tasks are executed and evaluated in the simulator (using the actor-critic) and the result updates the prompt.
% Exploration continues until the LLM decides that it has addressed all hints, the teacher intervenes, or a fixed iteration threshold is met.
We find that decomposing exploration to a chain of two LLM calls, prompted separately for compositions and variations, leads to faster completion. Task variations, proposed by the second LLM call, are not included in the progress prompt field. A temperature parameter of $0.1$ is set at successive iterations to encourage diverse responses.

% \subsection{Abstraction: Distilling Experiences into new Skills}
% \label{method:abs}
\noindent \textbf{Skill Abstraction} This module leverages LLMs' capabilities to define Python functions out of examples. The goal is dual: a) maintain the same code logic, but abstract code variations such as target objects and destination regions as arguments to the new function, and b) extract boilerplate code snippets and abstract them to new functions. This is achieved by prompting the LLM in two rounds. The prompt contains:
\begin{itemize}
    \item A \textbf{general purpose} system message that primes the LLM for function generation and imposes constraints.
    \item An \textbf{API} field, rendered from the dependencies of input code snippets as an import statement.
    \item A set of two \textbf{exemplar} function definitions. The exemplars are different for each round of abstraction. As in exploration, one exemplar is manual and the other is selected from the first LLM response of the last cycle. 
    \item The input code snippets with their instruction as comments. We first provide \textbf{definition examples}, which are the raw instruction-code pairs from each cluster of the experience memory, and then \textbf{abstraction examples}, which correspond to the LLM-generated functions from the first round.
\end{itemize}
We also ask the LLM to provide a docstring, which is used as the skill description, as well as to re-write the given examples based on the generated function, which is used to refactor the memory and move to the replay stage of the sleep phase.

\section{EXPERIMENTS}
The focus of our experimental evaluation is threefold: a) compare our method against previous baselines for tabletop manipulation in simulation (Sec.\ref{res:comparisons}), b) evaluate the impact of each our method's proposed contributions (Sec.\ref{res:ablations}), and c) demonstrate the transferability of our approach to the real world (Sec.\ref{res:sim2real}).

\subsection{Evaluation Setup}
\noindent \textbf{Implementation} We leverage OpenAI’s $\texttt{gpt-3.5-turbo}$ \cite{OpenAIChatGPT} engine for all LLM generations, and \texttt{text-embedding-ada-002} \cite{OpenAIEmbeddings} as the memory embedding model. Our system is built using the LangChain library \cite{Topsakal2023CreatingLL}.
Our simulator environment is built on Pybullet \cite{benelot2018} and it is based on the Ravens \cite{TransporterNR} manipulation suite, with the blocks-and-bowls setup replicated from previous works \cite{CodeAP, SocraticMC}. 
We introduce more tasks and language variations, for a total of 41 task templates, organized in a curriculum of 4 cycles: a) \textit{Spatial Coordination}, i.e. precise motions relative to objects/regions, b) \textit{Visual Reasoning}, i.e. determining attributes, resolving spatial relations and counting/enumerating objects, c) \textit{Object Manipulation}, i.e. single picking, releasing and placing tasks, and d) \textit{Rearrangement}, i.e. long-horizon tasks that involve multiple objects and destinations.

\noindent \textbf{Evaluation} For conducting generalization experiments, we generate task instances in three splits \cite{CodeAP}: seen instructions with either seen (SA) or unseen (UA) attributes, and unseen instructions with unseen attributes (UI). 
% We evaluate the agents in all three splits after each cycle and report averaged success rates. 
For studying learning in the lifelong setup, we also propose two more splits: a) a \textit{forward-transfer (FT)} split, which contains unseen compositions of tasks from the present cycle with all previous tasks (with seen attributes), and b) \textit{backward-transfer (BT)}, which contains the FT tasks from the previous cycle.
These splits are meant to study whether the agent can learn to transfer knowledge between tasks (FT), and to what extent it ``forgets" or improves on previous tasks (BT).

\noindent \textbf{Baselines} We consider four baselines: a) learning end-to-end multi-task policies with \textit{CLIPort} \cite{CLIPortWA}, adapted as in \cite{SocraticMC} (not applicable in all tasks), b) prompting LLMs for primitive-based policy code using a static prompt, as in \textit{CaP} \cite{CodeAP} (without hand-crafted routing between LLM sub-systems), c) \textit{LRLL-no-sleep}, where we remove the sleep phases from our \textit{LRLL} and only retrieve examples from the memory without abstraction, and d) \textit{LRLL-no-wake}, where we attempt to synthesize new skills directly from the human demonstrations, without the exploration of the wake phase. 

\begin{table}[t]
\caption{\footnotesize
    Averaged success rates (\%) over seen instructions with seen/unseen attributes (SA/UA) and unseen instructions (UI), organized in a 4-cycle curriculum with 10 trials per instruction. Best results are in bold.}
% \vspace{-2.5mm}
    \centering
    %begin{adjustbox}{width=.5\textwidth,center}
    
         \resizebox{\linewidth}{!}{%
    \begin{tabular}{lccccccccc}
    \toprule
   \multirow{2}{3em}{\textbf{Tasks}}  &  \multicolumn{3}{c}{\textbf{CLIPort} \cite{CLIPortWA}}  & \multicolumn{3}{c}{\textbf{LLM-static} \cite{CodeAP} } & \multicolumn{3}{c}{\textbf{LRLL} (ours) } \\ 
   \cmidrule(l{8pt}r{6pt}){2-4}
   \cmidrule(l{8pt}r{4pt}){5-7}
   \cmidrule(l{8pt}r{10pt}){8-10}
    & SA & UA & UI   & SA & UA & UI   & SA & UA & UI  \\
    \midrule
    Spatial Coord/on & - & - & - & \textbf{100.0} & \textbf{100.0}  & \textbf{100.0} & \textbf{100.0} & \textbf{100.0} & \textbf{100.0} \\
    Visual Reasoning & - & - & -& 90.0 & 83.3 & 66.6 & \textbf{91.7} & \textbf{94.0}  & \textbf{85.1}   \\
    Object Manip/on & \textbf{98.3} & 37.1 & 4.1 & 95.0 & 94.1 & 80.0   & 98.1 &  \textbf{98.9} &\textbf{90.4}   \\
    Rearrangement & 70.8 & 13.9 & 0.3 & 93.0 & 90.0 & 60.6  & \textbf{97.0}  & \textbf{95.4}  & \textbf{70.9}   \\
    \midrule
    Average &  84.5 & 25.5 & 2.2  & 94.5  & 91.9 & 76.8 & \textbf{96.9} & \textbf{97.1}  &  \textbf{86.6}  \\
     \bottomrule
    \end{tabular}%
    }
%\end{adjustbox}
    \label{tab:gen}
      \vspace{-1mm}
\end{table}
\subsection{Tabletop Manipulation in Simulation}
\label{res:comparisons}
We first wish to evaluate the performance of \textit{LRLL} compared to established baselines in our simulated tabletop domain. 
To that end, we developed a simulated teacher that samples tasks from a set of predefined templates.
The teacher generates up to 5 demonstration (1 per SA template) and multiple test (10 per UA, UI template) tasks in the beginning and end of each of our 4 cycles. 
For end-to-end learning with CLIPort \cite{CLIPortWA}, we sample $1k$ trajectories per task template using a scripted expect for each cycle and train the model incrementally.
For our LLM-static baseline \cite{CodeAP}, we append demos from each new cycle in the LLM's prompt. 
The same demos are provided to \textit{LRLL} at the beginning of each cycle's wake phase. 
Agents are tested at the end of each cycle. We repeat our experiments three times with different teacher seeds and report averaged success rates in Table~\ref{tab:gen}.

We observe that CLIPort struggles with unseen attributes and its performance degrades drastically with unseen instructions.
% Due to its pick-and-place based nature, CLIPort is also not applicable in simple coordination or visual reasoning tasks.
LLM-static is robust to unseen attributes ($2.9\%$ average drop) and can generalize significantly better in unseen task instructions, with an average success rate of $76.8$\% in all cycles. 
We find that this baseline's main limitation is producing non-executable code in cases of unseen instructions at later cycles, which we attribute to its inability to interpret and compose multiple skills from a limited demonstration context. 
% (tends to repeat existing code patterns from the demos or hallucinate unknown functions).
% \textit{prompt saturation} phenomena \cite{CodeAP, SocraticMC}
Such skill compositions are (partially) already explored during the wake phase of our \textit{LRLL}, and abstracted to functions during the sleep phase, resulting in policy code that is much shorter and functional in style. 
This robustifies \textit{LRLL}'s generated policies, which translates to an average increase of $\sim6\%$ in unseen attribute and $\sim10\%$ in unseen instructions compared to LLM-static.

 \subsection{Ablation Studies}
Our ablations focus on exploring the effect of each of our proposed components and discussing options for implementation.

\label{res:ablations}
\begin{table}[!t]
\caption{\footnotesize
    Averaged success rates  (\%) over unseen task combinations (FT) and a subset of previous task combinations (BT) for each cycle and baseline. Best results are in bold.}
% \vspace{-2.5mm}
    \centering
    %begin{adjustbox}{width=.5\textwidth,center}
    
         \resizebox{\linewidth}{!}{%
    \begin{tabular}{lcccccccc}
    \toprule
   \multirow{2}{3em}{\textbf{Tasks}}  &  \multicolumn{2}{c}{\textbf{LLM-static} \cite{CodeAP}}  & \multicolumn{2}{c}{\textbf{LRLL-no-wake} } & \multicolumn{2}{c}{\textbf{LRLL-no-sleep}} & \multicolumn{2}{c}{\textbf{LRLL}}\\ 
   \cmidrule(l{8pt}r{6pt}){2-3}
   \cmidrule(l{8pt}r{4pt}){4-5}
   \cmidrule(l{8pt}r{10pt}){6-7}
   \cmidrule(l{8pt}r{10pt}){8-9}
    & FT & BT & FT & BT & FT & BT & FT & BT   \\
    \midrule
    Spatial Coord/on & \textbf{100.0} & \textbf{100.0} & 68.7 & 60.0 & \textbf{100.0} & \textbf{100.0} & \textbf{100.0}& \textbf{100.0} \\
    Visual Reasoning & 60.0 & \textbf{100.0} & 40.0 & 55.0& \textbf{80.0} & \textbf{100.0} & \textbf{80.0}  & \textbf{100.0}\\
    Object Manip/on & 55.3 & 60.0 & 46.6 & 40.4 & 78.4 & 80.0 & \textbf{94.0} & \textbf{80.0}\\
    Rearrangement & 45.7   & 55.3 & 50.0 & 46.6 & 64.0 & 78.4 & \textbf{70.2} & \textbf{94.0} \\
    \midrule
    Average &  65.1 &  78.9  & 51.3 & 50.5 & 80.6 & 89.6 & \textbf{88.1} & \textbf{93.5}  \\
     \bottomrule
    \end{tabular}%
    }
%\end{adjustbox}
    \label{tab:ftbt}
      \vspace{-1mm}
\end{table}
% \begin{figure}[!t]
%     \includegraphics[width=\columnwidth]{img/ftbt.pdf} 
% \caption{\footnotesize
% Averaged success rates over unseen task combinations (left) and a subset of all previous tasks combinations (right) for each cycle.
% }
% \label{fig:ftbt}
%     \vspace{-3mm}
% \end{figure}

\noindent \textbf{Forward/Backward Transfer} We compare the averaged success rates of all baselines in FT/BT instructions. Results are reported in Table~\ref{tab:ftbt}. First, we assess that all baselines are robust in tasks from previous cycles, showcasing the immunity of LLM's in-context learning to forgetting. However, no actual increase in backward task's performance is reported in any baseline. For forward transfer, we see a large increase in averaged success between LLM-static and \textit{LRLL}. Even without refactoring code  (\textit{LRLL-no-sleep} baseline), retrieving explored experiences leads to better compositional abilities, with a $\sim15\%$ delta from static.   

\noindent \textbf{Static Prompts vs. Retrieval} When prompted with a few examples, the difference between a static and a retrieved prompt is marginal. In the late cycles of the curriculum, we observe the effect of \textit{prompt saturation} \cite{CodeAP, SocraticMC} kicking in the static baseline, leading to several instabilities in the LLM responses, such as ignoring the first examples in favour of more recent ones or referring to variable names outside the current scope. Retrieval-based baselines tackle such issues by ensuring a fixed context length for the LLM actor.

\begin{figure}[!t]
    \includegraphics[width=\columnwidth]{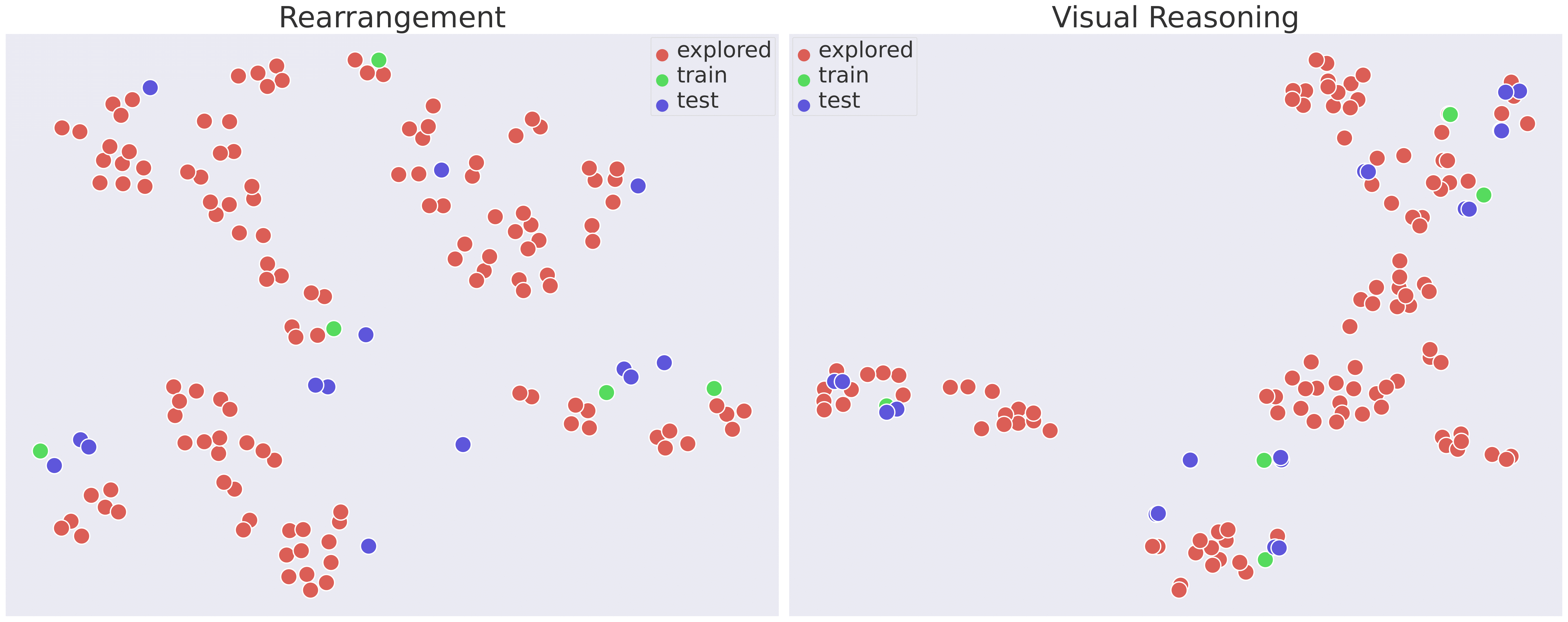} 
\caption{\footnotesize
\textit{tSNE} projections of train (SA), test (UA+UI) and explored task instruction embeddings, for two of our curriculum cycles: a) \textit{Visual Reasoning} (right), and b) \textit{Rearrangement} (left). The exploration module augments the agent's experiences with task variations that cover a broad range of skill compositions from the demos. \textit{(Best viewed in color)}.
}
\label{fig:tsne}
    \vspace{-3mm}
\end{figure}

\noindent \textbf{Effect of Exploration} The contribution of the exploration module is vital, as the performance of \textit{LRLL-no-wake} is consistently much lower across cycles. This is due to the high difficulty of abstracting skills from one-shot demos, which usually leads to a one-to-one mapping of instructions to functions, without adding any actual refactoring. When adding exploration, the abstractor has significantly more examples to define new skills. To evaluate the breadth of variance in the explored tasks, we visualize the \textit{tSNE} projections of their instruction embeddings \cite{OpenAIEmbeddings} compared to demonstration and test tasks within a cycle (see Fig.~\ref{fig:tsne}). 

\begin{figure}[!t]
    \includegraphics[width=\columnwidth]{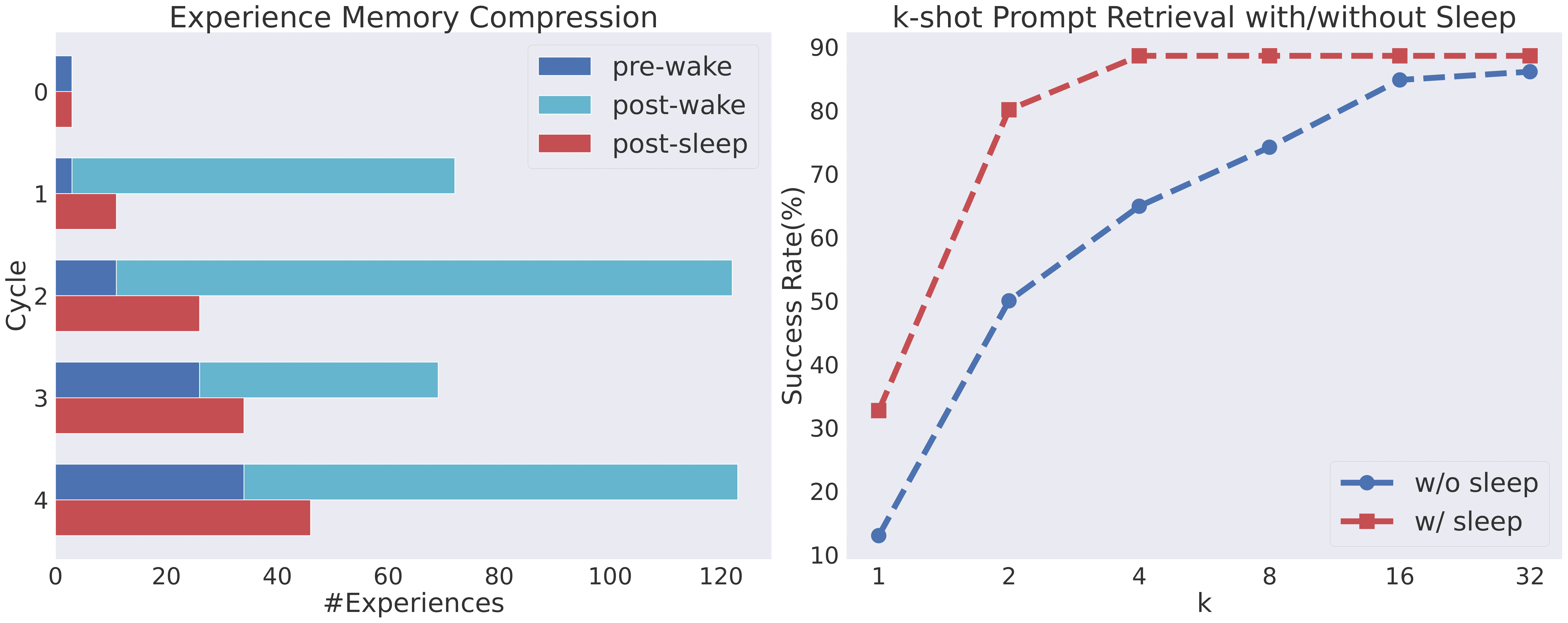} 
\caption{\footnotesize
(Left): Number of stored experiences per cycle before and after the sleep phase of \textit{LRLL}. Sleep helps to compress the number of experiences needed to reach the same performance. (Right): Averaged success rates in all unseen instructions vs. number of retrieved experiences. Sleep phase refactors experiences, which leads to sufficient context with fewer examples.
}
\label{fig:sleep}
    \vspace{-3mm}
\end{figure}

\noindent \textbf{Effect of Abstraction} \textit{LRLL-no-sleep} never abstracts the explored tasks into new skills, and so needs to retrieve a lot of examples in order to obtain sufficient context.
This effect bottlenecks the agent to the quality of the retriever.
% With the sleep phase, experiences are refactored to be just a call to the new skill, which allows providing sufficient context with few examples. 
Besides performance gains, sleep leads to other practical benefits (see Fig.~\ref{fig:sleep}). First, the amount of experiences required to maintain the same success within each cycle is drastically reduced, leading to a $\times 8$ decrease in RAM required to store experience embeddings. 
Second, \textit{LRLL} with refactored memory requires much less retrieved experiences to maintain high performance in UI tasks. 
Additionally, as the experiences are refactored to be simple function calls (to the abstracted skills), the retrieved code is itself smaller, which leads to smaller prompt lengths and hence cost gains for using GPT.

\noindent \textbf{LLM and Embedding Models} We find no significant difference between \texttt{text-davinci-003} and \texttt{gpt-3.5-turbo} in the quality of generated policy code or function abstraction. For exploration, we find that both models provide rich variance in the proposed tasks, but the chat model tends to be less responsive to the hints signal. This effect can be ameliorated by running more exploration iterations.
The choice of the embedding model is more important, as experiments with smaller encoder LMs such as Sentence-BERT \cite{Reimers2019SentenceBERTSE} showed a tendency to retrieve instructions that are similar lexically (e.g. same object noun appears), but not necessarily convey the same task. 
% Barplot with 5 cycles as x-axis and gen/on results in y-axis for three bars: lrll, lrll-no-sleep, lrll-no-wake.

% \begin{itemize}
%     \item effect of sleep-phase. Point to barplot 
%     \item efficiency of wake-phase exploration. Point to barplot. (+tSNE of actual and "imagined" tasks for some selected cycles).
%     \item (if time: effect of LLM size and type: Compare GPT with Llama-2-7b/13b/70b-quant)
% \end{itemize}

\subsection{Zero-Shot Sim-to-Real Transfer}
\label{res:sim2real}
We repeat our curriculum with \textit{LRLL} in a dual-arm robot setup with two UR5e arms and a Kinect sensor. We provide vision APIs for open-vocabulary detection with MDETR \cite{MDETRM} and attribute recognition with CLIP \cite{CLIP}.
To assist in articulated grasping, we also integrate GR-ConvNet \cite{Kumra2019AntipodalRG} for 4-DoF grasp synthesis as a vision API.
The motion primitives for moving the arm and opening/closing the fingers are parameterized by the left or right arm. 
We include a catalog of 12 household objects, including fruits, soda cans, juice boxes etc. 
We first train LRLL using our default 4-cycle curriculum in the Gazebo simulator \cite{Gazebo} and then test the agent in the real robot.
We demonstrate that the robot is able to perform long-horizon rearrangement tasks that combine precise spatial positioning with reasoning about object attributes, without any further adaptation from simulation.
% \footnote{A video of robot demonstrations is included as supplementary material.}
% A video of transfer demonstrations is included as supplementary material. 
Errors were observed mostly at motion execution due to collisions, as well as perception errors due to CLIP misclassifications.

% \begin{table}[t]
% \caption{\footnotesize
%     sth}
% % \vspace{-2.5mm}
%     \centering
%     %begin{adjustbox}{width=.5\textwidth,center}
    
%          \resizebox{\linewidth}{!}{%
%     \begin{tabular}{clccc}
%     \toprule
%     \textbf{Train/Test} & \textbf{Task Family} & \textbf{CLIPort} \cite{CLIPortWA} & \textbf{CaP} \cite{CodeAP} & \textbf{LRLL} \footnotesize{(ours)}\\
%     \midrule
%     SI/SA & VL-Reasoning &  88.4 &  &    \\
%     SI/SA &  Long-Horizon & 70.9  &  &     \\
%     \midrule
%     SI/UA & VL-Reasoning & 44.1  &  &      \\
%     SI/UA & Long-Horizon &  4.9 &  &    \\
%     \midrule
%     UI/UA & VL-Reasoning & 0.0  &  &      \\
%     UI/UA & Long-Horizon &  0.0 &  &    \\
%      \bottomrule
%     \end{tabular}%
%     }
% %\end{adjustbox}
%     \label{tab:gen}
%       \vspace{-1mm}
% \end{table}

\section{CONCLUSION \& LIMITATIONS}
In this work, we introduce \textit{LRLL}, an agent and learning algorithm for lifelong robot manipulation. 
\textit{LRLL} exploits the emergent capabilities of modern LLMs to: a) generate policies as code, b) interact with the environment to explore new tasks, and c) distill the acquired experiences into new skills over time.
\textit{LRLL} replaces tedious prompt engineering with retrieval from memory, and a static library of skills with an expandable codebase, written and verified by the agent itself.
Empirical evaluation shows that our agent learns a library of composable, generalizable, and interpretable skills that can be transferred to the real world, while
its dynamic and gradient-free nature prevents it from prompt saturation and forgetting phenomena of stationary-LLM and end-to-end approaches respectively.

\textit{LRLL} comes not without limitations. First, perception is restricted by the choice of vision APIs, which currently support only referring expressions and attribute classification. 
In the future, we would like to look at multimodal LLMs \cite{Instruct2ActMM, VisualIT} for open-ended vision-language grounding.
Second, the current human demonstration input to \textit{LRLL} is language, limiting its scalability to skills that can be expressed symbolically as primitive compositions.
For articulated, contact-rich manipulation tasks, we would like to augment demonstration input to support video or kinesthetic teaching.
Third, the initial prompts to exploration/abstraction modules need to be refined when changing domains or LLM engines. 
Finally, exploration with commercial LLMs is constrained by latency and price factors. In the future, we would like to investigate the gap between GPT and open-source alternatives \cite{Llama2O}.

\section{Acknowledgments}
We thank the Center for Information Technology of the University of Groningen for providing access to the Hábrók high-performance computing cluster.
\bibliography{main}

\end{document}